\documentclass[conference]{IEEEtran}
\IEEEoverridecommandlockouts
\usepackage{hyperref}
\usepackage{placeins}
\usepackage{multirow}
\usepackage{array}
\usepackage{cite}
\usepackage{booktabs}
\usepackage{amsmath,amssymb,amsfonts}
\usepackage{algorithmic}
\usepackage{graphicx}
\usepackage{textcomp}
\usepackage{xcolor}
\def\BibTeX{{\rm B\kern-.05em{\sc i\kern-.025em b}\kern-.08em
    T\kern-.1667em\lower.7ex\hbox{E}\kern-.125emX}}
\begin{document}

\title{DenoMAE: A Multimodal Autoencoder for Denoising Modulation Signals\\

\thanks{This work is funded by the U.S. Department of Energy, funding number DE-FE0032196.}
}

\author{\IEEEauthorblockN{1\textsuperscript{st} Atik Faysal}
\IEEEauthorblockA{\textit{Electrical and Computer Engineering} \\
\textit{Rowan University}\\
Glassboro, USA \\
faysal24@rowan.edu}
\and
\IEEEauthorblockN{2\textsuperscript{nd} Taha Boushine}
\IEEEauthorblockA{\textit{Electrical and Computer Engineering} \\
\textit{Rowan University}\\
Glassboro, USA \\
bouhsi95@rowan.edu}
\and
\IEEEauthorblockN{3\textsuperscript{rd} Mohammad Rostami}
\IEEEauthorblockA{\textit{Electrical and Computer Engineering} \\
\textit{Rowan University}\\
Glassboro, USA \\
rostami23@rowan.edu}
\and
\IEEEauthorblockN{4\textsuperscript{th} Reihaneh Gh. Roshan}
\IEEEauthorblockA{\textit{Computer Science} \\
\textit{Stevens Institute of Technology}\\
Hoboken, NJ \\
rghasemi@stevens.edu}
\and
\IEEEauthorblockN{5\textsuperscript{th} Huaxia Wang}
\IEEEauthorblockA{\textit{Electrical and Computer Engineering} \\
\textit{Rowan University}\\
Glassboro, USA \\
wanghu@rowan.edu}
\and
\IEEEauthorblockN{6\textsuperscript{th} Nikhil Muralidhar}
\IEEEauthorblockA{\textit{Computer Science} \\
\textit{Stevens Institute of Technology}\\
Hoboken, NJ \\
nmurali1@stevens.edu}

\and
\IEEEauthorblockN{7\textsuperscript{th} Avimanyu Sahoo}
\IEEEauthorblockA{\textit{Electrical and Computer Engineering} \\
\textit{ University of Alabama in Huntsville}\\
Huntsville, AL \\
avimanyu.sahoo@uah.edu}

\and
\IEEEauthorblockN{8\textsuperscript{th} Yu-Dong Yao}
\IEEEauthorblockA{\textit{Electrical and Computer Engineering} \\
\textit{Stevens Institute of Technology}\\
Hoboken, NJ \\
yyao@stevens.edu}
}

\maketitle

\begin{abstract}

We propose Denoising Masked Autoencoder (DenoMAE), a novel multimodal autoencoder framework for denoising modulation signals during pretraining. DenoMAE extends the concept of masked autoencoders by incorporating multiple input modalities, including noise as an explicit modality, to enhance cross-modal learning and improve denoising performance. The network is pre-trained using unlabeled noisy modulation signals and constellation diagrams, effectively learning to reconstruct their equivalent noiseless signals and diagrams. DenoMAE achieves state-of-the-art accuracy in automatic modulation classification tasks with significantly fewer training samples, demonstrating a $10\times$ reduction in unlabeled pretraining data and a $3\times$ reduction in labeled fine-tuning data compared to existing approaches. Moreover, our model exhibits robust performance across varying signal-to-noise ratios (SNRs) and supports extrapolation on unseen lower SNRs. The results indicate that DenoMAE is an efficient, flexible, and data-efficient solution for denoising and classifying modulation signals in challenging, noise-intensive environments. Our codes are public at \href{https://github.com/atik666/denoMae/tree/master}{GitHubDenoMAE}.

\end{abstract}


\begin{figure*}[htbp]
    \centering
    \includegraphics[width=0.9\linewidth]{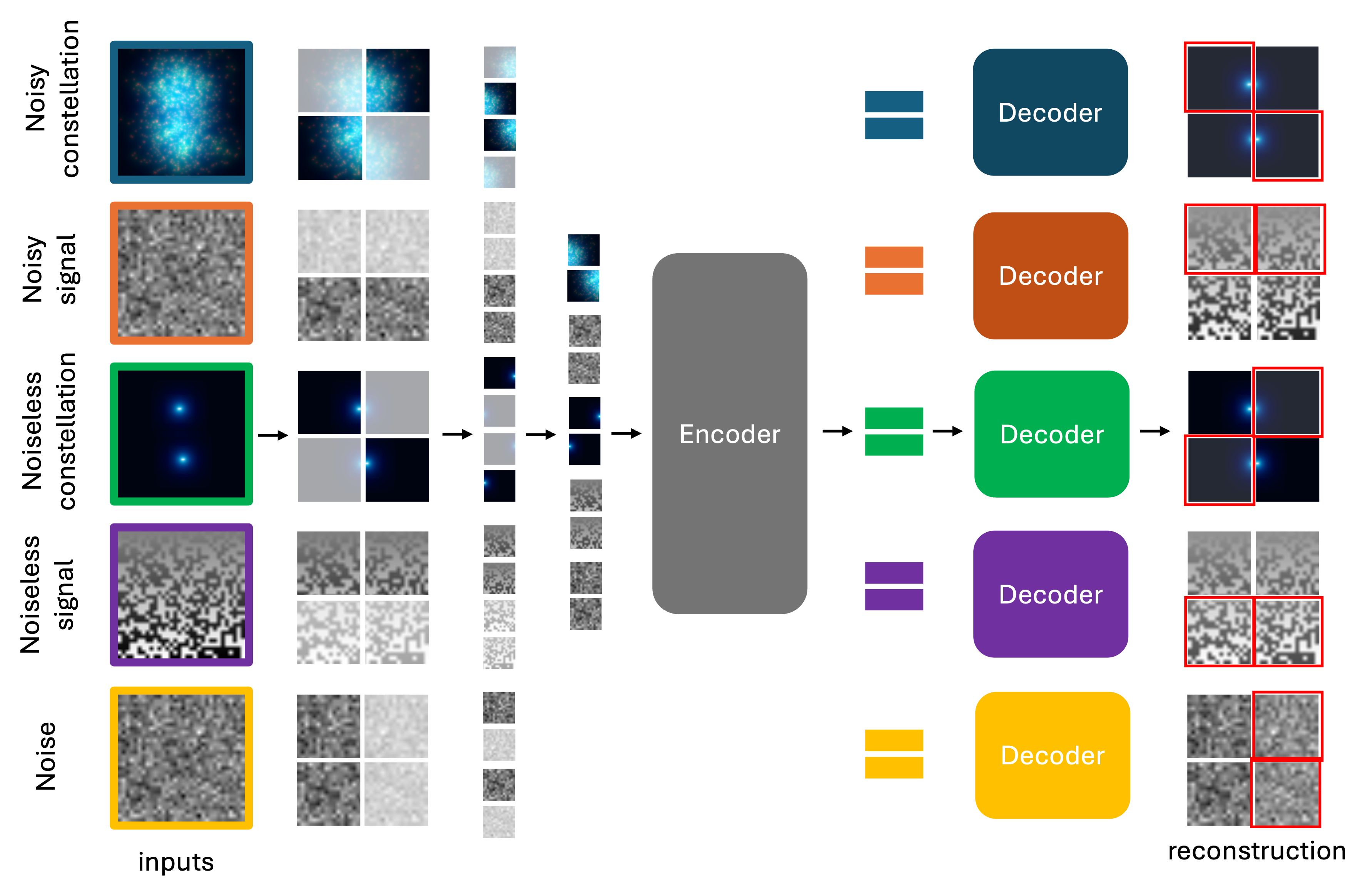} 
    \caption{
    \textbf{DenoMAE Pretraining Strategy}: We apply a random 75\% masking (not to scale in the illustration) across all input modalities. The remaining 25\% of visible patches are processed by a shared encoder, while each modality utilizes a dedicated decoder to reconstruct its masked patches. Only the encoder is reused for fine-tuning the downstream tasks.
    }
    \label{fig:architecture} 
\end{figure*}

\section{Introduction}

In recent years, the deep learning paradigm has shifted to address the high costs and scarcity of labeled data and the intensive data demands of supervised models. Self-supervised learning has emerged as an efficient solution, allowing models to leverage abundant raw data in pretraining. This approach drastically reduces the need for labeled data during fine-tuning while maintaining strong downstream performance \cite{balestriero2023cookbookselfsupervisedlearning}. Self-supervised methods have shown remarkable success across fields like natural language processing \cite{roberts2019exploring} and computer vision \cite{caron2021emerging, radford2021learning}, paving the way for new applications in fields with sparsely labeled data, including communications.

In communication systems, particularly wireless networks, maintaining signal integrity is crucial, yet it’s often compromised by limited data availability and varying noise levels during transmission. Estimating modulation signals becomes especially challenging as noise levels fluctuate with changing SNRs \cite{zheng2024deep}. This limitation becomes particularly problematic when models encounter previously unseen modulation types \cite{9965389}. While existing models can often classify known modulation schemes, they struggle to generalize to new or noisy conditions, resulting in misclassification and necessitating extensive retraining \cite{perenda2021learning, yang2019adaptive}. This problem highlights an urgent need for robust, adaptive models that can generalize across diverse conditions with minimal reliance on labeled data.

One approach to improving generalizability and robustness is through pretraining methods that not only capture robust representations but also denoise signals effectively, enhancing classification accuracy in downstream tasks and providing scalable solutions in dynamic communication environments. Masking-based methods have traditionally been effective for denoising and improving signal clarity, but they are often limited to single-modality approaches. Though effective, these methods are inherently constrained by the information within their specific modality, which restricts their potential to leverage cross-modal interactions for more robust learning \cite{shi2023gaf, huang2022deep}. Recent advances in multimodal learning address this limitation by integrating data from multiple modalities, enhancing model robustness and enabling more flexible zero-shot learning capabilities. In works like \cite{bachmann2022multimae, bachmann20244m}, attention mechanisms and masked modeling pretraining across modalities have successfully reconstructed masked signals, proving beneficial in improving classification under complex conditions \cite{cohen2023joint, leiber2022dipencoder}.

Building on recent advancements, we adapt these techniques and present DenoMAE, a multimodal autoencoder tailored for the unique challenges in communication systems. DenoMAE's innovative approach involves treating noise as an additional modality, allowing the model to generate all other modalities from a single input modality. This unique framework helps the network gain a comprehensive understanding of noise across modalities, significantly enhancing its denoising capabilities. A random masking strategy, which hides 75\% of the input modalities, is employed to challenge the model in reconstructing the missing patches, reducing computational complexity, and introducing beneficial uncertainty conducive to effective learning.

Our contributions to DenoMAE are threefold:

\begin{itemize} 
\item We propose a novel multimodal paradigm that incorporates noise as an additional modality, enabling more effective pretraining on modulation signals and enhancing model resilience to noise. 
\item We demonstrate that our pre-trained model achieves significant improvements in classification accuracy under low SNR conditions, with accuracy gains of up to 22.1\% over non-pre-trained models. 
\item Our approach outperforms traditional methods, achieving superior results while requiring only one-tenth of the unlabeled data for pretraining and one-third of the labeled data for fine-tuning, underscoring its efficiency and scalability for real-world applications. 
\end{itemize}

\section{Related Work}

\textbf{Modulation Classification}

Recent advancements in deep learning have transformed Automatic Modulation Classification (AMC) by enabling advanced analysis of constellation and other image diagrams. Much of the research emphasizes CNN-based architectures designed to classify modulation types while effectively handling uncertainties introduced by varying channel conditions \cite{zheng2019fusion, hermawan2020cnn, peng2017modulation, huynh2020mcnet}. Attention-based models like LSTMs and GRUs have been used widely for processing modulation signals as time-series data \cite{8446021, cheng2024automatic, zhang2022research, liu2021automatic}. More recently, transformer-based architectures have been adopted to leverage self-attention for improved classification accuracy \cite{kong2023transformer, kong2021transformer, cai2022signal}. Despite these advancements, these architectures often lack robustness across diverse noise scenarios, limiting their generalization beyond specific tasks.



\textbf{Multisensory Masked Modeling}

Masking large portions of the input and training the model to reconstruct or predict missing elements has become a popular self-supervised approach, especially within autoencoder-based methods. For example, BERT \cite{devlin2018bert} introduced masked language modeling by masking words in sentences and training the model to predict them, a concept adapted for vision with Masked Autoencoders (MAE), where image patches are masked to encourage the model to learn robust visual representations \cite{he2021mae}.

In multimodal contexts, masked modeling extends to tasks combining text, vision, and audio, promoting cross-modal understanding. For instance, Vision-Language Pretraining (VLP) models \cite{lu2019vilbert, chen2020uniter, wang2022image} utilize masked language and vision modeling to jointly learn from masked image regions and text tokens. Additionally, image modality-centered models like MultiMAE \cite{bachmann2022multimae}, 4M \cite{mizrahi20244m}, and 4M21 \cite{bachmann20244m} have demonstrated strong performance by pretraining on multiple image modalities and enhancing inference capabilities across diverse inputs. These multimodal masking methods have demonstrated significant gains in self-supervised learning, advancing model performance in tasks that benefit from multimodal reasoning.

\section{Methodology}
In this work, we propose DenoMAE, an innovative pretraining method for modulation classification that leverages recent advancements in multisensory pretraining. By incorporating noise as an additional modality during pretraining, DenoMAE significantly reduces data requirements while enhancing performance in downstream classification tasks.

\begin{figure}[htbp]
    \centering
    \includegraphics[width=0.9\linewidth]{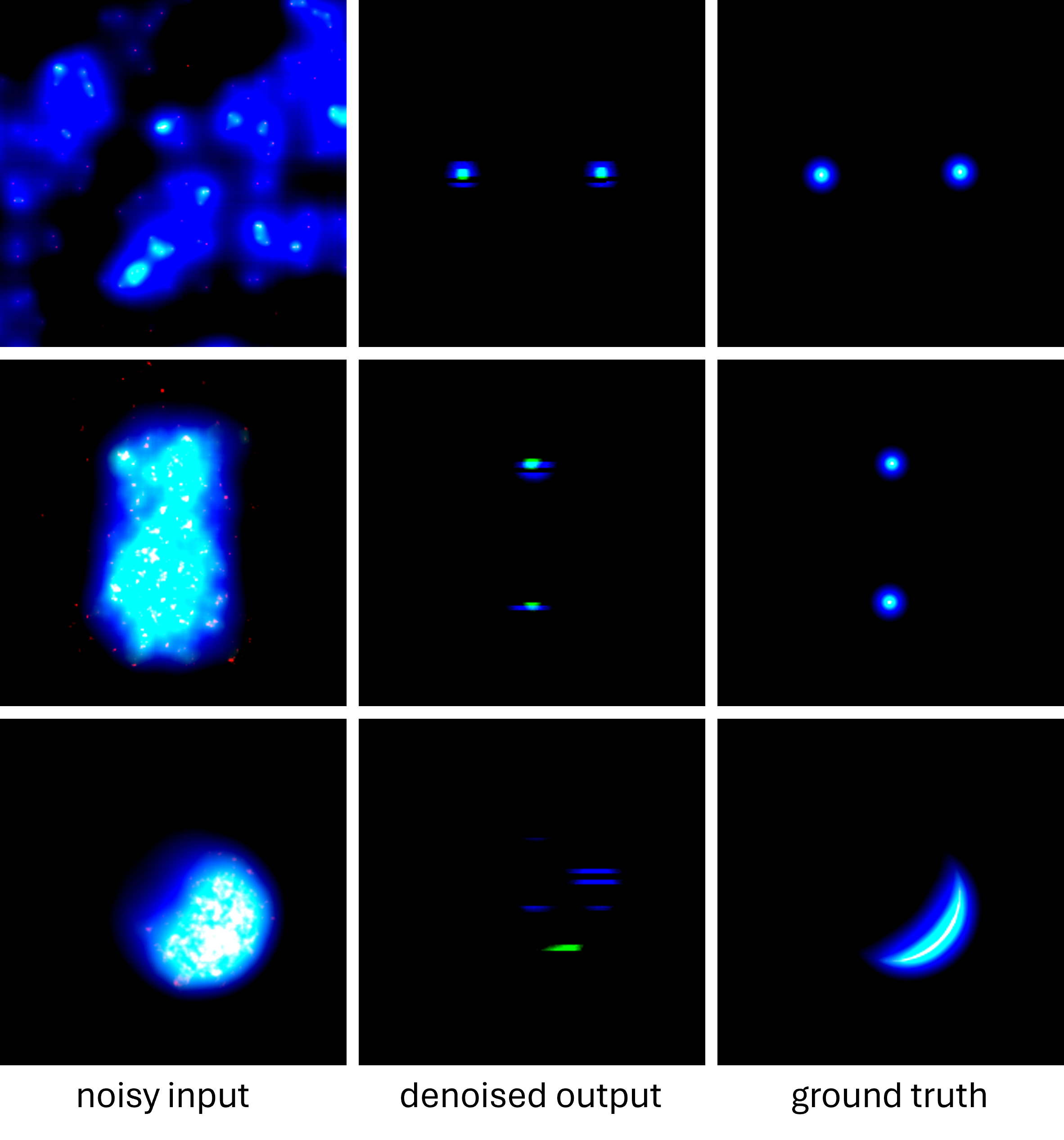} 
    \caption{Denoised outputs of DenoMAE on unlabeled constellation diagrams at different SNRs during pretraining.}
    \label{fig:outputs} 
\end{figure}

\subsection{DenoMAE Framework}

DenoMAE consists of four primary components: input embedding, encoder, shared latent space, and decoder. Figure \ref{fig:architecture} illustrates the overall architecture.

\subsubsection{Input Embedding}

For each modality $m \in {1, ..., n}$, we represent the input as a C-channel image $X_m \in \mathbb{R}^{C \times H \times W}$. We divide each image into non-overlapping patches of size $p \times p$, resulting in $N = (H/p)^2$ patches per modality. Each patch is flattened and linearly projected to a $D$-dimensional embedding space:
$$E_m = [e^m_1, ..., e^m_N] + E_{pos}, \quad e^m_i \in \mathbb{R}^D$$
where $E_{pos}$ represents learnable positional embeddings.

\subsubsection{Encoder}

The encoder $f_{enc}$ is a series of $L$ transformer blocks. We randomly mask a high proportion (e.g., 75\%) of the input patches for each modality. The encoder processes the visible patches:
$$H_m = f_{enc}([e^m_{i_1}, ..., e^m_{i_k}]), \quad i_j \in \text{visible patches}$$

\subsubsection{Shared Latent Space}

To enable cross-modal learning, we project the encoded representations from all modalities into a shared latent space. We use a linear projection followed by layer normalization:
$$Z_m = \text{LN}(W_m H_m + b_m)$$
The shared latent representation $Z$ is the concatenation of all modality-specific projections:
$$Z = [Z_1; ...; Z_5]$$

\subsubsection{Decoder}

The decoder $f_{dec}$ reconstructs the masked patches for all modalities simultaneously. It takes the shared latent representation $Z$ as input and processes it through a series of transformer blocks. The output is then split into modality-specific representations and projected back to the original patch space:
$$\hat{X}m = f{proj_m}(f_{dec}(Z))$$
where $f_{proj_m}$ is a modality-specific projection layer.

\subsubsection{Loss Function}

We employ a multi-modal reconstruction loss that combines individual losses for each modality:
$$\mathcal{L} = \sum_{m=1}^5 w_m \mathcal{L}_m$$
where $w_m$ is the weight for modality $m$, and $\mathcal{L}_m$ is the reconstruction loss for that modality. We use the mean squared error (MSE) as the reconstruction loss:
$$\mathcal{L}m = \frac{1}{|\mathcal{M}m|} \sum{i \in \mathcal{M}m} | \hat{X}{m,i} - X{m,i} |_2^2$$
\section{Experimental Setup}
We begin this section by detailing the parameters used for sample generation, followed by the classifier's configuration. Finally, we present the parameters applied to our downstream tasks.

\subsection{Constellation Diagram Generation}

Constellation diagrams offer a more informative approach to modulation classification compared to raw time-series signals, capturing richer detail essential for accurate interpretation \cite{doan2020learning, 9289413}. The process of generating constellation diagrams for modulation signal classification involves multiple stages. We begin by mapping modulated signals onto a $7 \times 7$ complex plane, which provides sufficient space to capture signal samples while maintaining computational efficiency for SNR ranges of -10 dB to 10 dB. The basic constellation diagram is then enhanced through a multi-step process: first, as a gray image that handles varying pixel densities where multiple samples may occupy single pixels; second, as an enhanced grayscale image that employs an exponential decay model ($B_{i,j}$) to account for both the precise position of samples within pixels and their influence on neighboring pixels. This model considers the sample point's power ($P$), the distance between sample points and pixel centroids ($d_{i,j}$), and an exponential decay rate ($\alpha$) \cite{peng2018modulation}. To make the representation compatible with DenoMAE, which expects RGB input, we generate a three-channel image by creating three distinct enhanced grayscale images from the same data samples, each utilizing different exponential decay rates. 

\subsection{Parameters in Sample Generation} 

The dataset comprises signals from ten modulation formats, each originally of length $L_0 = 1024$. To conform to the transformer's input dimensions, signals undergo a two-step preprocessing: (1) reshaping to $\mathbf{S}_1 \in \mathbb{R}^{32 \times 32}$, and (2) interpolation to $\mathbf{S}2 \in \mathbb{R}^{224 \times 224}$. For formats with $L_0 > 1024$ (e.g., $L_\text{GMSK} = 8196$), signals are initially downsampled to $L_0$. To match the three-channel structure of constellation images $\mathbf{I} \in \mathbb{R}^{3 \times 224 \times 224}$, $\mathbf{S}_2$ is replicated across three channels, yielding $\mathbf{S}_3 \in \mathbb{R}^{3 \times 224 \times 224}$. All signals are sampled at $f_s = 200$ kHz, ensuring consistency across modulation formats.

\begin{figure}[htbp]
    \centering
    \includegraphics[width=0.9\linewidth]{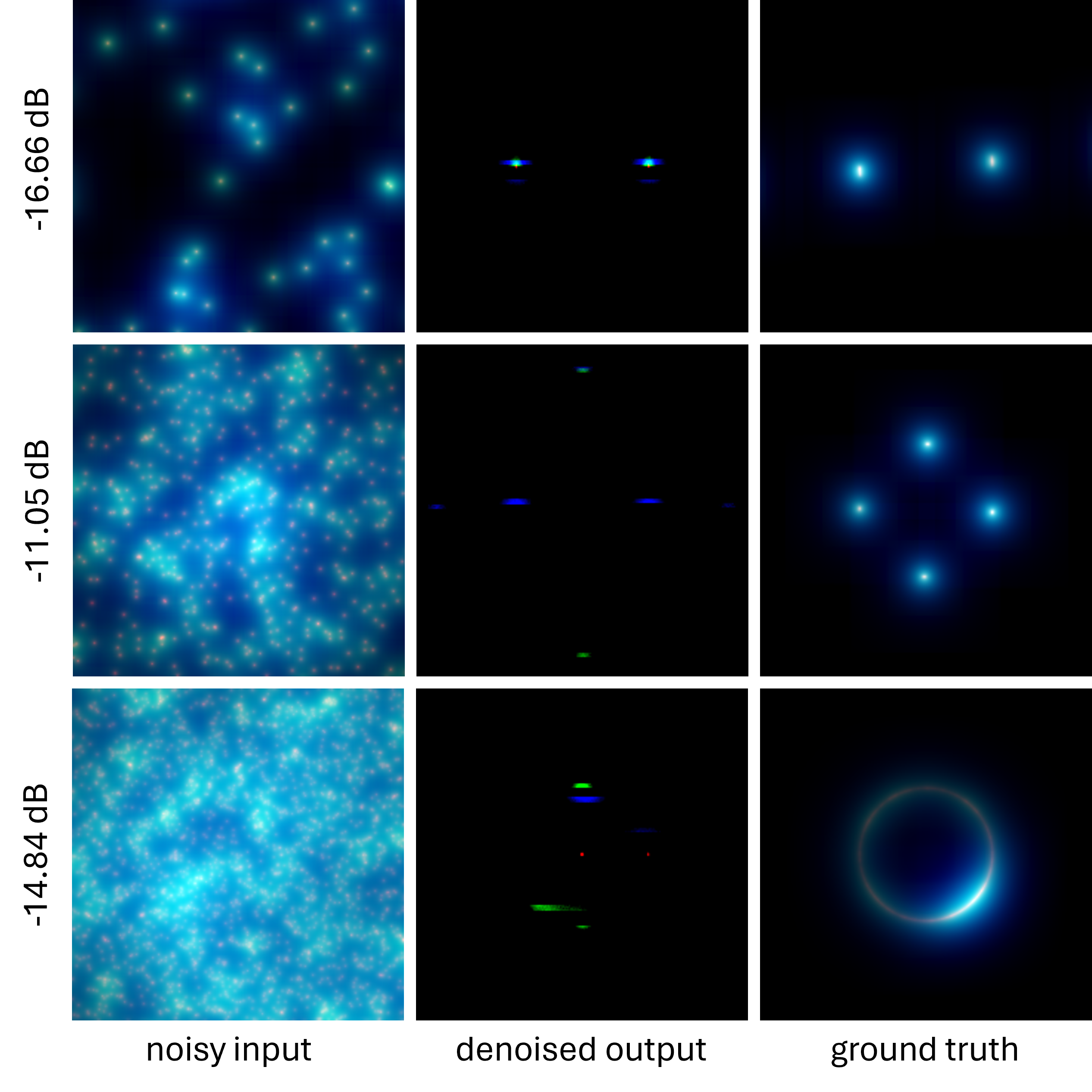} 
    \caption{Extrapolation-ability of DenoMAE on out-of-bound much lower SNRs.}
    \label{fig:extrap} 
\end{figure}

\subsection{Parameters for DenoMAE}

The MultiMAE architecture processes $n$ =5 modalities of input images $\mathbf{X} \in \mathbb{R}^{3 \times 224 \times 224}$, segmented into $16 \times 16$ patches. The model employs a transformer-based encoder-decoder structure with $d_\text{model} = 768$ embedding dimensions. The encoder consists of $L_e = 12$ layers, while the decoder uses $L_d = 4$ layers, both utilizing $h = 12$ attention heads. During training, a random masking strategy with $p_\text{mask} = 0.75$ is applied. The model is optimized using AdamW with learning rate $\eta = 10^{-4}$ over $E = 100$ epochs, using batches of size $B = 64$. This flexible architecture allows for easy adaptation to various multi-modal denoising tasks.

\subsection{Finetuning Parameters}

The fine-tuning task adapts the pre-trained DenoMAE model for classification across $C = 10$ classes. The encoder $E(\cdot)$ from DenoMAE, with $d_\text{model} = 768$, is used as a backbone, followed by a classifier head $H(\cdot)$ consisting of a two-layer MLP with hidden dimension 512 and dropout $p = 0.5$. For an input $\mathbf{X} \in \mathbb{R}^{3 \times 224 \times 224}$, the model computes $\hat{y} = H(G(E(\mathbf{X})))$, where $G(\cdot)$ is global average pooling. The model is trained for $E = 150$ epochs using Adam optimizer with learning rate $\eta = 10^{-4}$ and batch size $B = 32$. Training data undergoes normalization but no augmentation.
\section{Results and Discussion}



\begin{table}
\centering
\caption{Performance Comparison Across Modalities in Downstream Task}
\label{modulations}
\begin{tabular}{c|c|c|c|c|c}
\hline
\multicolumn{2}{c|}{Constellation} & \multicolumn{2}{c|}{Signal} & Noise & Test Accuracy (\%) \\
\cline{1-4}
Clean & Noisy & Clean & Noisy &  &  \\
\hline
\cline{1-4} 
\checkmark &  &  &  &  & 81.30 \\
\checkmark & \checkmark &  &  &  & 81.90 \\
\checkmark & \checkmark & \checkmark &  &  & 83.20 \\
\checkmark & \checkmark & \checkmark & \checkmark & & 83.30 \\
\checkmark & \checkmark & \checkmark & \checkmark & \checkmark & 83.50 \\
\hline
\end{tabular}
\end{table}


This work centers on denoising and reconstructing constellation diagrams, a modality used exclusively during downstream classification tasks. In the following sections, we provide detailed insights into those findings.

\begin{figure}[htbp]
    \centering
    \includegraphics[width=0.9\linewidth]{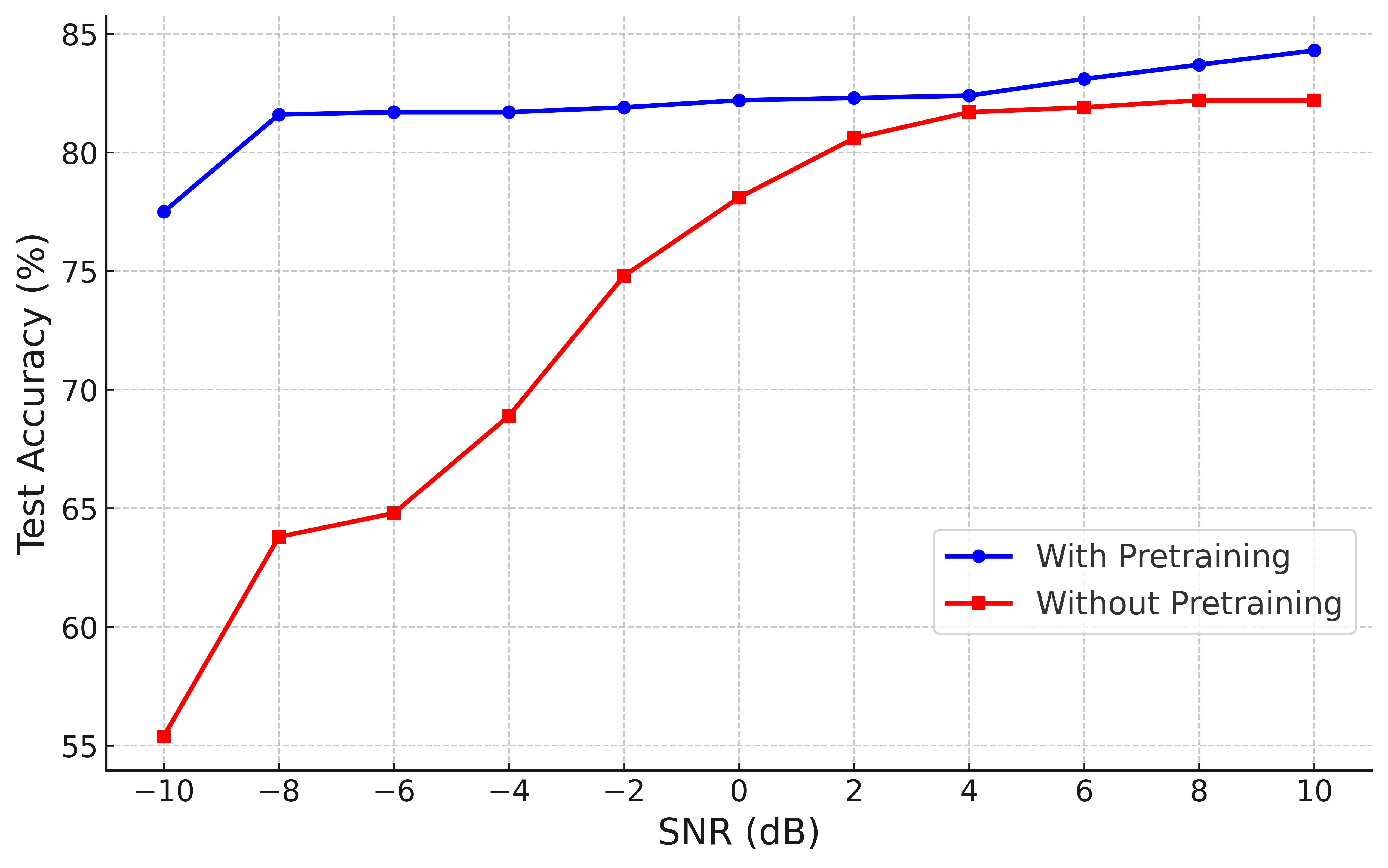} 
    \caption{DenoMAE fine-tuned classification accuracy for constellation diagrams at different SNRs.}
    \label{accuracy} 
\end{figure}

\begin{table*}[htbp]
\centering
\caption{Comparison of DenoMAE with other modulation classification methods}
\label{compare}
\begin{tabular}{cccccc}
\toprule
\multirow{2}{*}{\textbf{Methods}} & \multicolumn{2}{c}{\textbf{Number of samples}} & \multirow{2}{*}{\textbf{SNR dB}} & \multirow{2}{*}{\textbf{Number of classes}} & \multirow{2}{*}{\textbf{Test accuracy (\%)}} \\ 
\cmidrule(lr){2-3}
 & \textbf{Pretraining} & \textbf{Fine-tuning} & & \\ \midrule

AlexNet \cite{peng2018modulation}  & 800,000 (train) & 8,000 (test)  & 2 & 8 & 82.8  \\
NMformer & 106,800 & 3,000 & 0.5 - 4.5 & 10 & 71.6 \\
CNN-AMC \cite{meng2018automatic} & 79,200 & 228,060 & -6 & 4 & 50.40 \\
DL-GRF \cite{sun2022automatic} &  2,000 (train) & 200 (test) & 0 & 4 & 59 \\
DenoMAE (\textbf{Ours}) & 10,000 & 1,000 & 0.5 & 10 & 83.5 \\
DenoMAE (\textbf{Ours}) & 10,000 & 1,000 & -10 & 10 & 77.50 \\

\bottomrule
\end{tabular}
\end{table*}

\subsection{Pretraining Performance}

The pretraining effectiveness of DenoMAE was evaluated using 10,000 unlabeled signal samples across five modalities. Visual inspection of the denoised constellation diagram outputs (Figure \ref{fig:outputs}) demonstrates strong reconstruction capabilities, particularly for samples in the first two rows where the model successfully removes noise while preserving signal characteristics. While reconstruction quality slightly diminishes for samples in the last row, the overall shape and pattern retention remain sufficient for downstream tasks.

\subsection{Impact of SNR}

In fine-tuning, we used 1000 labeled constellation images with 10 classes. The impact of SNR on classification performance reveals DenoMAE's robust noise resilience. As shown in Figure \ref{accuracy}, the pre-trained model achieves 77.50\% accuracy even at -10 dB SNR, with performance steadily improving to 84.30\% at 10 dB. The pretraining advantage is most pronounced in challenging low-SNR conditions, demonstrating a 22.1\% improvement over the non-pre-trained baseline at -10 dB. This performance gap gradually narrows at higher SNRs, suggesting that pretraining is particularly valuable for learning noise-robust features.


\subsection{Extrapolation on out-of-bound SNRs}

DenoMAE demonstrates impressive extrapolation capabilities when evaluated on SNR conditions significantly outside its training range. Although trained only on SNR levels above -10 dB, the model continues to deliver meaningful reconstruction performance at challenging SNR levels between -11 dB and -20 dB. As shown in Figure \ref{fig:extrap}, DenoMAE effectively denoises highly noisy input constellation diagrams, making the first two samples visually recognizable. The final sample, while impacted by extreme noise, still approximates the outer shape of the ground truth.

\subsection{Ablation Study on Modality Impact}

Our ablation study on modality impact, detailed in Table \ref{modulations}, reveals the effectiveness of the multimodal approach. The full model with five modalities (noisy image, noisy signal, noiseless image, noiseless signal, and noise) achieves 83.50\% accuracy. Performance degrades gracefully as modalities are removed, with even the single-modality version maintaining 81.30\% accuracy. This suggests that while additional modalities contribute to performance gains, the model maintains robust performance even with reduced modality input, indicating efficient cross-modal learning capabilities.

\subsection{Comparative Analysis}


In Table \ref{compare}, we demonstrate the superior performance of DenoMAE against existing modulation classification approaches through comprehensive empirical evaluation. Our method achieves state-of-the-art accuracy of 83.5\% using only 10,000 pretraining and 1,000 fine-tuning samples, representing a dramatic reduction in data requirements compared to previous methods (e.g., 80× fewer samples than AlexNet, 10.7× fewer than NMformer). Notably, DenoMAE maintains robust performance across challenging SNR conditions, achieving 77.50\% accuracy even at -10 dB SNR, while handling a broader set of 10 modulation classes compared to methods like CNN-AMC and DL-GRF which only classify 4 classes. This performance significantly outpaces existing approaches, with improvements of up to 33.1\% over CNN-AMC (50.40\%) and 24.5\% over DL-GRF (59\%) under comparable or more challenging conditions, demonstrating DenoMAE's exceptional efficiency and generalization capabilities in modulation classification tasks.
\section{Conclusion}

In this work, we introduced DenoMAE, a multimodal autoencoder designed for the denoising of modulation signals. Our method leverages the benefits of a masked autoencoder framework while incorporating multiple modalities, including noise, to enhance the model’s robustness and efficiency. Through extensive experiments, we demonstrated that DenoMAE achieves state-of-the-art performance with significantly reduced pretraining and labeled data requirements compared to existing methods. The ablation studies further highlighted the model’s flexibility in utilizing various modalities, maintaining high classification accuracy even with reduced modality input. Our results suggest that DenoMAE not only excels in traditional denoising and classification tasks but also holds the potential for extrapolation on out-of-bound low SNRs, broadening its applicability in real-world communications environments. Future work could explore the extension of DenoMAE to more diverse signal types and further optimize its architecture for lower computational overhead.


\FloatBarrier
\bibliographystyle{ieeetr}
\bibliography{ref}

\end{document}